\documentclass[letterpaper, 10 pt, conference]{ieeeconf}  

\IEEEoverridecommandlockouts                              

\usepackage{graphics} 
\usepackage{epsfig} 
\usepackage{mathptmx} 
\usepackage{times} 
\usepackage{amsmath} 
\usepackage{amssymb}  
\usepackage{balance}
\usepackage{booktabs}
\usepackage{multirow}
\usepackage{gensymb}
\usepackage[pagebackref,breaklinks,colorlinks]{hyperref}
\usepackage{cleveref}
\usepackage{xspace}
\let\labelindent\relax
\usepackage{enumitem}

\title{\LARGE \bf
Benchmarking Classical and Learning-Based \\Multibeam Point Cloud Registration
}

\author{Li Ling$^{1}$, Jun Zhang$^{2}$, Nils Bore$^{3}$, John Folkesson$^{1}$, Anna Wåhlin$^{4}$
\thanks{This work was supported by Stiftelsen för Strategisk Forskning (SSF) through the Swedish Maritime Robotics Centre (SMaRC) (IRC15-0046). The computations were enabled by the supercomputing resource Berzelius provided by National Supercomputer Centre at Linköping University and the Knut and Alice Wallenberg Foundation, Sweden. The data was collected as part of the TARSAN project during the International Thwaites Glacier Collaboration (ITGC) cruise in 2022. The expedition was supported by The Swedish Research Council under grant 2021-03679\_VR. We would like to thank the AUV technicians Anders Sjövall and Mark Symons for the enormous help in data collection. We are grateful for expert support from the captain and crew onboard RVIB \textit{Nathaniel B Palmer}. Expert logistic support was provided by NSF-U.S. Antarctic Program and NERC-British Antarctic Survey (BAS).}
\thanks{$^{1}$Division of Robotics, Perception and Learning (RPL), KTH Royal Institute of Technology, Stockholm, Sweden
        {\tt\small \{liling,johnf\}@kth.se}}%
\thanks{$^{2}$Institute of Computer Graphics and Vision (ICGV), TU Graz, Austria
        {\tt\small jun.zhang@tugraz.at}}%
\thanks{$^{3}$Ocean Infinity, Sven Källfelts Gata 11, SE-426 71 Västra Frölunda, Sweden
        {\tt\small nils.bore@oceaninfinity.com}}%
\thanks{$^{4}$Department of Marine Sciences, University of Gothenburg, Sweden
        {\tt\small anna.wahlin@gu.se}}%
}

\begin{document}

\makeatletter
\DeclareRobustCommand\onedot{\futurelet\@let@token\@onedot}
\def\@onedot{\ifx\@let@token.\else.\null\fi\xspace}

\def\eg{\emph{e.g}\onedot} \def\Eg{\emph{E.g}\onedot}
\def\ie{\emph{i.e}\onedot} \def\Ie{\emph{I.e}\onedot}
\def\cf{\emph{c.f}\onedot} \def\Cf{\emph{C.f}\onedot}
\def\etc{\emph{etc}\onedot} \def\vs{\emph{vs}\onedot}
\def\wrt{w.r.t\onedot} \def\dof{d.o.f\onedot}
\def\etal{\emph{et al}\onedot}
\makeatother

\newcommand{\mbesdataset}{\textit{DotsonEast Dataset}}

\maketitle
\thispagestyle{empty}
\pagestyle{empty}

\begin{abstract}
Deep learning has shown promising results for multiple 3D point cloud registration datasets. However, in the underwater domain, most registration of multibeam echo-sounder (MBES) point cloud data are still performed using classical methods in the iterative closest point (ICP) family. In this work, we curate and release \mbesdataset, a semi-synthetic MBES registration dataset constructed from an autonomous underwater vehicle in West Antarctica. Using this dataset, we systematically benchmark the performance of 2 classical and 4 learning-based methods. The experimental results show that the learning-based methods work well for coarse alignment, and are better at recovering rough transforms consistently at high overlap (20-50\%). In comparison, GICP (a variant of ICP) performs well for fine alignment and is better across all metrics at extremely low overlap (10\%). To the best of our knowledge, this is the first work to benchmark both learning-based and classical registration methods on an AUV-based MBES dataset. To facilitate future research, both the code and data are made available online.\footnote{The code and data are available at \href{https://github.com/luxiya01/mbes-registration-data/}{https://github.com/luxiya01/mbes-registration-data/}}
\end{abstract}

\section{Introduction}
\label{sec:sec1-introduction}
Multibeam echo-sounder (MBES) is the de-facto sensor for underwater surveys performed by surface vessels, autonomous underwater vehicles (AUVs) and remote operating vehicles (ROVs) \cite{micallefSubmarineGeomorphology2018}. When mounted on AUVs, MBES can provide high resolution bathymetry data at inaccessible or dangerous locations, such as under glaciers \cite{graham2022rapid}. 
Since MBES provides bathymetry as 3D point cloud, this sensor is also useful for AUV localization \cite{romanSelfConsistentBathymetricMapping2007, torrobaAutonomousIndustrialscaleBathymetric2019}.
For MBES-based AUV localization frameworks, loop closure detection and registration of MBES submaps is an integral component \cite{torrobaComparisonSubmapRegistration2018, tanDatadrivenLoopClosure2022}.

Most existing MBES registration methods are based on the iterative closest point (ICP) family, with errors measured using the root-mean-square (RMS) consistency error metric introduced by Roman and Singh \cite{romanConsistencyBasedError2006}. Despite the success of ICP algorithms in MBES registration, this family of algorithm is more suited for final adjustment of roughly aligned point clouds, due to their sensitivity to initial transformations \cite{brossardNewApproach3D2020}. Since many MBES registration methods use the dead-reckoning pose as the initial alignment \cite{romanSelfConsistentBathymetricMapping2007, torrobaAutonomousIndustrialscaleBathymetric2019}, a large dead-reckoning error will lead to erroneous registrations.

\begin{figure}
    \centering
    \includegraphics[width=\linewidth]{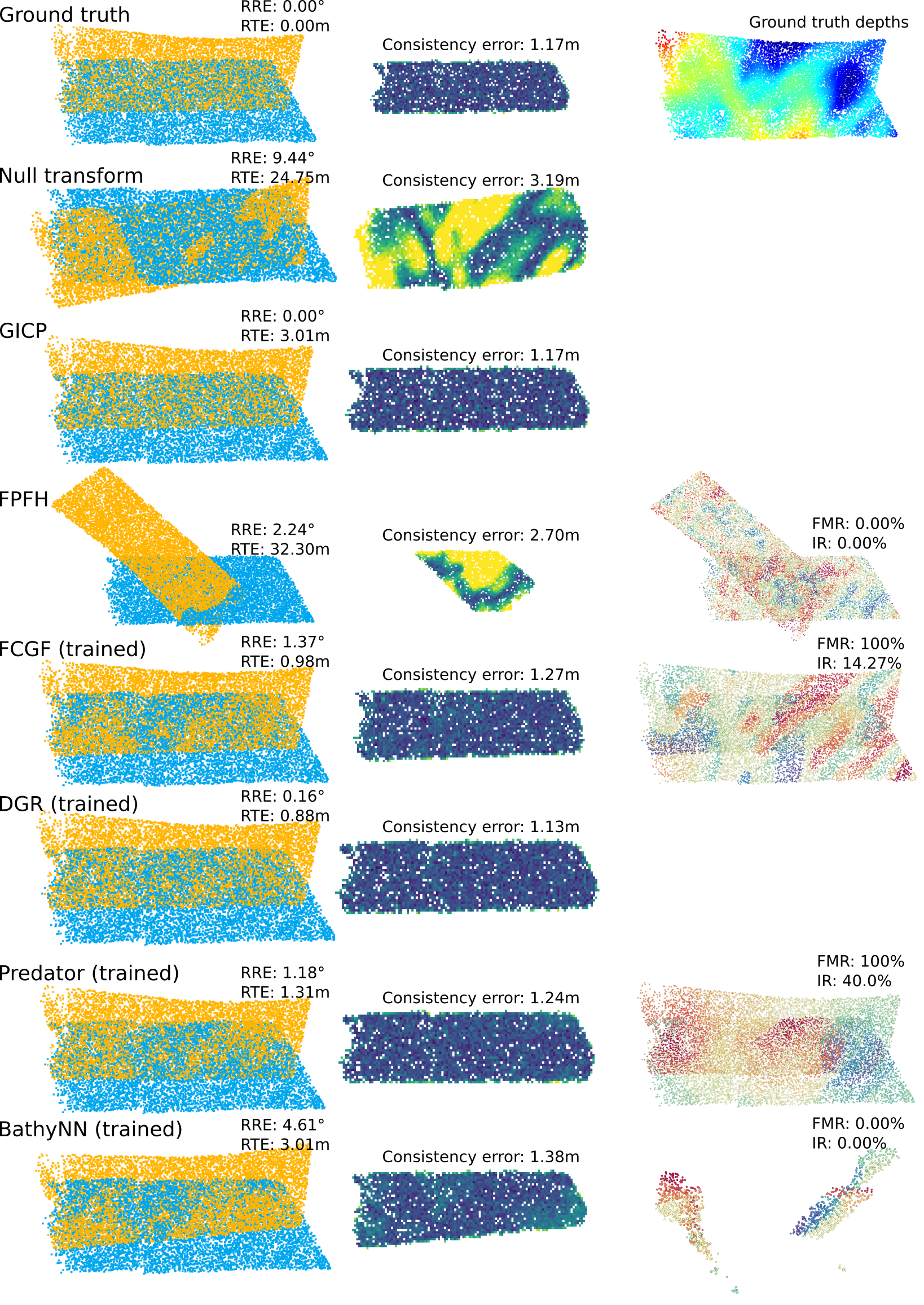}
    \caption{Example MBES submap pair from the proposed \mbesdataset~. Each row showcases the predicted transformations (\textbf{left}), the consistency error of the map (\textbf{middle}) and the t-SNE \cite{van2008visualizing} embedding of the feature descriptors (\textbf{right}). For methods without feature descriptors, the right column is left blank. The ground truth and null transforms are provided for comparison. The right column on the ground truth row shows the point cloud pair colored by depth. More details on the dataset and its metrics can be found in \autoref{sec:sec3-multibeam-benchmark}.}
    \label{fig:example-data}
\end{figure}

In recent years, learning-based approaches have shown promising results for a variety of 3D point cloud datasets, including simulated CAD models \cite{wu3dShapenetsDeep2015}, outdoor LiDAR data \cite{geigerAreWeReady2012} and RGBD indoor datasets \cite{zeng3dmatchLearningLocal2017}. However, the application of learning based methods to bathymetric point clouds is lacking, and so is a comprehensive dataset and framework for systematic evaluation of various registration methods. This can be attributed to two factors. Firstly, underwater MBES data is costly to collect and often viewed as sensitive security information of a nation. As such, the availability of open sourced data is severely restricted. Secondly, underwater datasets often lack ground truth, making the evaluation of registration results difficult. This is particularly true for MBES data, which requires a \textit{moving} platform to collect. As such, the groundtruth of each MBES swath depends on the instantaneous pose. Due to this evaluation difficulty, most MBES registration work rely on the RMS error for quantitative evaluation and qualitatively demonstrate specific features on the seafloor to showcase the improvement in consistency \cite{barkbyFeaturelessApproachEfficient2011, romanSelfConsistentBathymetricMapping2007, torrobaAutonomousIndustrialscaleBathymetric2019}.


In this work, we attempt to bridge the aformentioned gap by curating a large MBES dataset from field data collected by \href{https://www.gu.se/en/skagerak/auv-autonomous-underwater-vehicle}{RAN}, Gothenburg University's Hugin AUV in West Antarctica during the \href{https://thwaitesglacier.org/}{International Thwaites Glacier Collaboration} research cruise in 2022. From the raw data, a semi-synthetic registration dataset is generated by sampling random transformations to create pairs of submaps. This semi-synthetic workflow allows us to obtain ground truth transformations, which is often unavailable during real AUV bathymetry missions. Using these ground truth labels, we train multiple state-of-the-art deep learning models for MBES registration, and evaluate both classical and learning-based models using a diverse set of metrics. \autoref{fig:example-data} showcases an example of MBES pair from our dataset, as well as the predicted registration using the evaluated methods. The data and the proposed benchmark metrics are available \href{https://github.com/luxiya01/mbes-registration-data/}{here}.

In summary, the contributions of this paper is as follows:
\begin{enumerate}[leftmargin=.5cm]
    \item We propose a method to construct a semi-synthetic registration dataset from raw MBES survey data, and a framework to systematically evaluate registration methods.
    \item We open source \mbesdataset, a large scale semi-synthetic MBES point cloud registration dataset from AUV field data collected in front of Dotson Ice Shelf in West Antarctica.
    \item We evaluate multiple state-of-the-art registration methods on our dataset, including both classical and learning-based methods, and highlight the unique challenges of MBES registration.
\end{enumerate}
To the best of our knowledge, this is the first published work to benchmark both classical and learning-based methods specifically for MBES point clouds.

\section{Related Work}
\label{sec:sec2-related-work}
This section starts with a brief summary of existing families of point cloud registration methods. We then delve deeper into the registration methods for MBES point clouds. Finally, we highlight the importance of open source datasets for algorithmic developments in point cloud registration.

\subsection{Point Cloud Registration} \label{sec:sec2-subsec-point-cloud-registration}
Point cloud registration refers to the task of recovering the real-world transformation between two partially overlapping point clouds with unknown correspondences \cite{yewRpmnetRobustPoint2020}. Existing methods can be divided into feature-free and feature-based methods, both with hand-crafted and learning based variants. Whilst feature-free methods return a transformation directly, feature-based methods require outlier rejection methods to estimate a final robust transformation. More comprehensive surveys can be found in \cite{huangComprehensiveSurveyPoint2021} and \cite{zhangDeepLearningBased2020}.

\subsubsection{Feature-Free Methods}
The feature-free methods provide an estimate of rigid transformation directly from the point cloud, without the extra step of feature correspondence matching. Traditionally, iterative closest point (ICP) \cite{beslMethodRegistration3D1992} is the de-facto algorithm for direct registration. However, the success of ICP relies on good initial alignment and relatively low noise point clouds with few outlier correspondences, limiting its usefulness to fine alignment of point clouds with initial coarse alignment provided by other methods. There is an abundance of research addressing the short comings of ICP, see \cite{pomerleauReviewPointCloud2015} for a more comprehensive review. 

\subsubsection{Feature-Based Methods}
Feature based methods generally follows a two-step process. In the first step, descriptive and robust feature descriptors are generated from both point clouds. Correspondences of these descriptors are then computed and a robust estimator such as RANSAC \cite{fischlerRandomSampleConsensus1981} is used to estimate a rigid transformation. Many handcrafted descriptors have been designed for 3D point clouds, most of which try to summarize local surface signatures such as point distribution \cite{johnsonUsingSpinImages1999}, surface curvatures \cite{rusuFastPointFeature2009} and normals \cite{tombariUniqueSignaturesHistograms2010} into histograms. A more comprehensive survey on handcrafted features is found in \cite{guoComprehensivePerformanceEvaluation2016}.

In recent years, learning based methods have surpassed handcrafted descriptors in multiple benchmarks such as ModelNet \cite{wu3dShapenetsDeep2015}, 3DMatch \cite{zeng3dmatchLearningLocal2017} and KITTI \cite{geigerAreWeReady2012} datasets. These methods can roughly be divided keypoint free dense methods and keypoint-based methods. The keypoint free methods compute descriptors on the entire (often downsampled) point cloud. Notably, the fully convolutional geometric feature (FCGF) \cite{choyFullyConvolutionalGeometric2019} is the first method that densely extract point cloud features using sparse convolution \cite{choy4dSpatiotemporalConvnets2019} based fully convolutional networks.

The keypoint-based methods aim to extract distinguishable, repeatable and matchable keypoints from 3D point clouds. For instance, Predator \cite{huangPredatorRegistration3d2021} focuses on detecting matchable keypoints in overlapping regions, with a two-stream encoder-decoder network coupled with a graph neural network based overlap attention module for inter-stream information propagation, allowing for successful registration of point cloud pairs with low overlap ratio ($< 30\%$).

\subsubsection{Outlier Rejection Methods}
RANSAC \cite{fischlerRandomSampleConsensus1981} is the most commonly used outlier rejection method for robust estimation of point cloud registration, and has been demonstrated to work well with both classical and learning-based feature descriptors. However, RANSAC suffers from slow convergence and low accuracy at high outlier ratios, motivating the development of  robust estimators that have higher tolerance to outliers \cite{zhouFastGlobalRegistration2016, yangTeaserFastCertifiable2020b}, as well as learning-based inlier/outlier rejection methods such as Deep Global Registration (DGR) \cite{choyDeepGlobalRegistration2020, baiPointdscRobustPoint2021, pais3dregnetDeepNeural2020}.


\subsection{Multibeam Point Cloud Registration}
\label{sec:sec2-subsec-mbes-registration}
As briefly mentioned in \autoref{sec:sec1-introduction}, the registration of multibeam bathymetric point cloud has challenges unique to the sensor domain. Firstly, bathymetric data often lacks well-defined landmarks such as man-made structures \cite{torrobaComparisonSubmapRegistration2018}, \cite{romanSelfConsistentBathymetricMapping2007}, making detection of repeatable keypoints difficult. Secondly, when mounted on AUVs, the multibeam data lacks registration ground truth, making the results difficult to evaluate. As such, the multibeam registration problem is normally presented as part of a AUV localization or SLAM problem, and seldom taken apart to be studied on its own. Most existing work focus on handcrafted methods, with many employing the feature-free ICP family algorithms \cite{romanSelfConsistentBathymetricMapping2007, palomerMultibeam3DUnderwater2016, torrobaAutonomousIndustrialscaleBathymetric2019}, and a few studies utilizing feature based methods \cite{hammondAutomatedPointCloud2015, sureshActiveSLAMUsing2020}. Some studies also model the terrain directly using Gaussian processes \cite{bore2018sparse, torroba2023online}. Amongst the ICP family methods, Torroba \etal \cite{torrobaComparisonSubmapRegistration2018} found generalized ICP (GICP) to achieve the most consistent result for bathymetric point cloud registration.



Although learning based methods have largely replaced handcrafted methods in other point cloud registration tasks, the usage of learning based methods in bathymetric point cloud registration is lacking. The most notable work is by Tan \etal \cite{tanDatadrivenLoopClosure2022}, where a neural network was designed and trained for keypoint selection and feature extraction of multibeam point clouds, and evaluated on loop closure and coarse registration tasks using data from an AUV mission in Thwaites Glacier in West Antarctica.

\subsection{Point Cloud Registration Datasets}\label{sec:sec2-subsec-point-cloud-registration-datasets}
The existence of structured, open source datasets allows for easier benchmarking of algorithms and has been beneficial for communities such as computer vision \cite{dengImagenetLargescaleHierarchical2009}. Early learning-based methods focused on the ModelNet40 \cite{wu3dShapenetsDeep2015} dataset, consisting of synthetically generated CAD models of 40 object categories. In recent years, however, this synthetic dataset is found to be overly simplistic. To prove their generalizability and real world usability, newer methods also evaluate on 3DMatch \cite{zeng3dmatchLearningLocal2017} - an indoor RGBD dataset and KITTI \cite{geigerAreWeReady2012} - an outdoor Lidar scan dataset.



The \mbesdataset~we propose here is a semi-synthetic one. By using real bathymetric point clouds collected by an AUV, we maintain aspects such as noisy measurements and non-homogeneous point distributions that exist in real underwater surveys. However, in order to obtain ground truth registration, we crop point clouds and synthesize rigid transformations using a procedure inspired by Yew and Lee \cite{yewRpmnetRobustPoint2020} (See details in \autoref{sec:sec3-sub-multibeam-dataset}). A natural future extension to this work would be to replace the synthesized transformation with real ones using the recently released multibeam datasets with GPS ground truth \cite{krasnoskyBathymetricMappingSLAM2022}, \cite{bernardiAURORAMultisensorDataset2022}.

\section{Multibeam Benchmark}
\label{sec:sec3-multibeam-benchmark}
\subsection{Raw Data Description and Processing}
\begin{figure}[t]
  \centering
   \includegraphics[width=\linewidth]{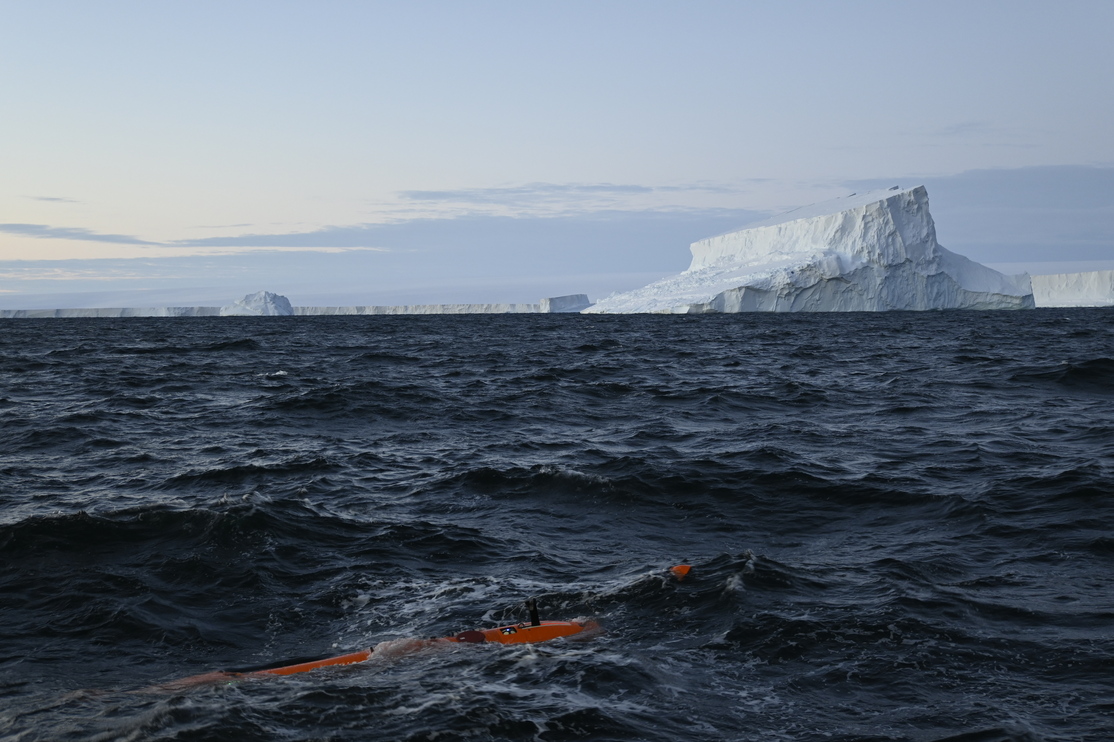}
   \caption{Kongsberg's Hugin AUV (orange) during the recovery of a survey in West Antarctica.}
   \label{fig:data-collection-site}
\end{figure}

\begin{figure}[t]
    \centering
    \includegraphics[width=.49\textwidth]{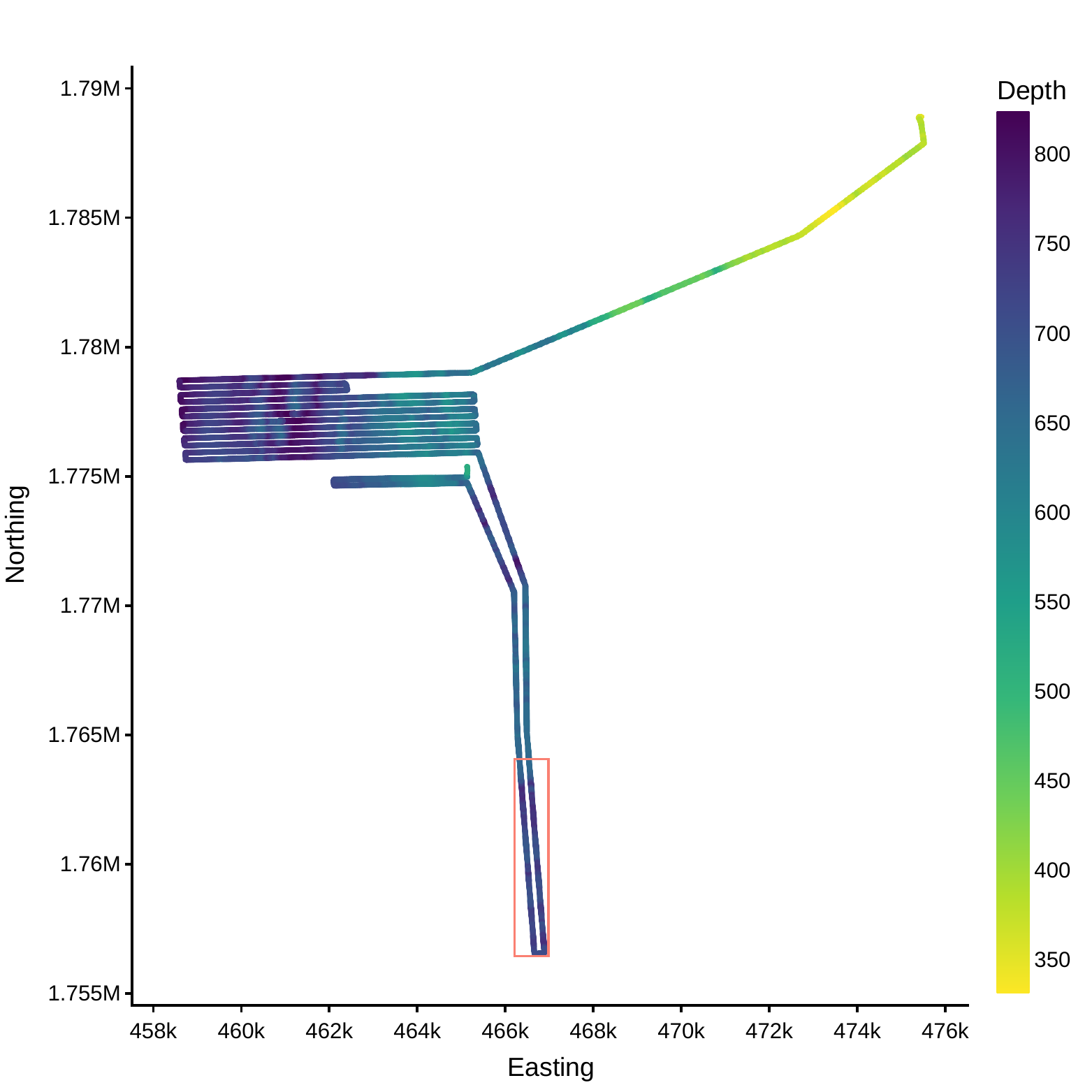}
    \caption{AUV trajectory throughout the mission, as reported by the onboard INS system. The color of the trajectory represents the vehicle depth. The red rectangle highlights the test data segment, which is chosen to be physically furthest away from the training data.}
    \label{fig:auv-trajectory}
\end{figure}

The data presented in this paper was collected using \href{https://www.gu.se/en/skagerak/auv-autonomous-underwater-vehicle}{RAN} - Gothenburg University's Kongsberg Hugin AUV equipped with a Kongsberg EM2040 multibeam echo sounder during the 2022 \href{https://thwaitesglacier.org/}{ITGC} cruise. The survey site was close to the eastern side of Dotson ice shelf in West Antarctica (see \autoref{fig:data-collection-site}). The vehicle trajectory estimated by the onboard inertial navigation system (INS) is shown in \autoref{fig:auv-trajectory}. Throughout most of the survey, the AUV was set to keep an altitude of $100 m$ above the seafloor. The sound velocity recorded by the multibeam receiver is used for data processing. \autoref{tab:survey-details} presents the details of the survey and sensor settings.

The original data was stored in Kongsberg's \textit{.all} format. The offshore EIVA software \href{https://www.eiva.com/products/navisuite/navisuite-processing-software/naviedit-pro}{NaviEdit Pro} was used to synchronize multibeam measurements with the INS positioning system and to convert the data into the correct UTM coordinate system. 
The exported files were then converted into numpy arrays with the dimension of $no.pings \times no.beams \times 3$ (x,y,z hits of each beam) to allow for easier manipulation. In order to retain the noisy measurements typical in a bathymetric survey, no data cleaning was performed. 

\begin{table}[h!]
  \centering
  \caption{Survey details and sensor settings}
  \label{tab:survey-details}
  \begin{tabular}{@{}lc@{}}
    \toprule
    Details & Specifications \\
    \midrule
    Vehicle speed & 2 m/s \\
    Vehicle altitude & $\sim$ 100 m \\
    Survey length & $\sim$ 138 km \\
    Survey duration & $\sim$ 19 h \\
    Sonar frequency & 400 kHz \\
    Ping rate & 2.5 Hz ($\sim$ 0.4 s/ping) \\
    Beam forming & 400 beams across 120\degree \\
    Total number of pings & 188 844 pings\\
    \bottomrule
  \end{tabular}
\end{table}

\subsection{Multibeam Dataset}
\label{sec:sec3-sub-multibeam-dataset}
To construct a registration dataset from the collected data, short sections of consecutive MBES pings are grouped together into submaps \cite{torrobaComparisonSubmapRegistration2018, bichucherBathymetricFactorGraph2015}.
In submap-based multibeam SLAM problems, the submap size is chosen so that each submap contains enough geometric variation for registration, whilst being small enough to be viewed as drift-free \cite{romanSelfConsistentBathymetricMapping2007,torrobaAutonomousIndustrialscaleBathymetric2019}. In this work, we set each submap to contain 100 consecutive pings. To maximize data usage, the step size is set to 20 pings, i.e. two consecutive submaps contain 80\% data overlap. This overlap is by design, as it allows for the construction of submap pairs with 20\%, 40\%, 60\%, 80\% and 100\% overlap, facilitating a more diverse range of evaluations. Using this schema, a total of 9415 submaps are constructed. The dataset is further split into train/val/test subsets, each with 7263/1206/946 submaps. 
The test set location is highlighted by the orange box in \autoref{fig:auv-trajectory}. The large lawnmower survey data collected before the orange box is used as the the training set, and the small section after the orange box serves as the validation set.
This dataset split is chosen to maximize the physical distance between the training and test subsets in order to prevent data leakage. For all experiments, we use $1m$ voxel downsampling to achieve uniform point density, then randomly sample maximum $10k$ points for each submap. 

To convert the submaps to suit registration tasks, we take inspiration from a similar procedure outlined by \cite{wangDeepClosestPoint2019}, \cite{yewRpmnetRobustPoint2020} and \cite{tanDatadrivenLoopClosure2022}, but modify the sampled transformations to better suit our application domain. Specifically, given a pair of multibeam submaps $(X, Y)$, we first crop each of the submaps independently to obtain $(X_c, Y_c)$. Noting the 2.5D nature of bathymetric point cloud, we sample a random vector in the XY-plane instead of in the full 3D space, and shift this vector to crop the submap such that approximately 70\% of the submap is retained. This means that after cropping, the effective overlap ratio between $(X_c, Y_c)$ is the $0.7 \times 0.7 \approx 50\%$ of the original overlap between $(X, Y)$. For instance, an original submap pair with $20\%$ overlap has an effective overlap ratio of $10\%$ after cropping. 

To synthesize submap pairs for registration, we then rotate, translate and add noise to each submap individually. For rotation, we sample Z-axis rotations in the range of [0, 10\degree], whilst keeping the XY axis rotation intact. This is because in AUV missions, the vehicle heading error is bounded by the INS system, and rotation errors around XY axes are mostly negligible. For translation, we sample the X- and Y-axis translation in the range [-40, 40] meters independently, simulating the potential dead reckoning drift in XY directions. For Z-axis, we sample a much smaller range of [-2, 2] meters, since the Z values are normally provided by pressure sensors and are less prone to accumulated drifts. The sampled rotation and translation are combined into a rigid transformation matrix $\mathbf{T}$, and the task is to register the transformed source submap $X_c = \mathbf{T}Y_c$ to the original reference submap $Y_c$. \autoref{fig:example-data} provides an example of the final submap pair used for registration.

\subsection{Evaluation Metrics}
Compared to previous work on multibeam registration, such as \cite{romanSelfConsistentBathymetricMapping2007, torrobaAutonomousIndustrialscaleBathymetric2019, hammondAutomatedPointCloud2015, bichucherBathymetricFactorGraph2015} and \cite{tanDatadrivenLoopClosure2022}, the proposed \textit{DotsonEast Dataset} benefits from having the sampled ground truth transformations, allowing for a more complete set of evaluation metrics. We propose evaluating the registration results using three sets of metrics, including the bathymetric specific consistency error, transformation accuracy metrics and feature correspondence metrics for feature-based methods.

\textbf{Consistency error} introduced by Roman and Singh \cite{romanConsistencyBasedError2006} is a point-based error metric designed to estimate bathymetric surface thickness and highlight wrongly registered regions in point cloud data. 
This error metric circumvents the problem of lack of ground truth transformation. As such, 
it is frequently used in real AUV surveys to measure the effectiveness of SLAM algorithms and resulting map quality \cite{torrobaComparisonSubmapRegistration2018, torrobaAutonomousIndustrialscaleBathymetric2019, tanDatadrivenLoopClosure2022}. Computing consistency error requires gridding the overlapping point clouds into a joint grid. For our dataset, the resolution of the grid is set to $2m$, twice the resolution of the point cloud. This resolution is chosen to retain enough details in the gridded map, whilst keeping the number of unoccupied grids low. The consistency error is sometimes hard to evaluate, since a small value can caused by both an accurate ground truth transformation and a transformation prediction that keep the two submaps almost disjoint. To aid the analysis, we also report the \textit{predicted overlap (\%)}, defined as the percentage of grid cells receiving hits from both submaps. To simulate real-life AUV missions where ground truth transformation is unknown, consistency metrics are computed for all successfully returned transformations, regardless of their accuracy. The accuracy of transformations are evaluated using metrics described below. A method might fail to return a transformation at all due to failure of convergence, too few correspondences, etc.

\textbf{Transformation accuracy metrics} include \textit{relative rotation error (RRE)}, \textit{relative translation error (RTE)} and \textit{registration recall (RR)}. They directly evaluate the accuracy of predicted transformations and have been widely applied to the KITTI dataset \cite{choyFullyConvolutionalGeometric2019, baiD3featJointLearning2020, choyDeepGlobalRegistration2020}. In our dataset, a transformation is considered successfully recalled with $RRE \leq 5 \degree$ and $RTE \leq 10 m$.

\textbf{Feature correspondence metrics} measure the quality of suggested feature matches and are thus only applicable to feature-based methods such as FPFH \cite{rusuFastPointFeature2009} and FCGF \cite{choyFullyConvolutionalGeometric2019}. Following standard metrics for the indoor 3DMatch dataset \cite{zeng3dmatchLearningLocal2017}, we report the \textit{feature match ratio (FMR)}, defined as the percentage submap pairs that have enough ($\geq 5\%$) \textit{inlier} matches under ground truth transformation, where inlier threshold is set to $2 m$, twice the resolution of the downsampled point clouds. The intermediate \textit{inlier ratio (IR)} for FMR computation is also recorded.

\section{Experiments}
\label{sec:sec4-experiments}

\subsection{Evaluated Methods}
We evaluate state-of-the-art classical and learning-based methods from each family of methods detailed in \autoref{sec:sec2-subsec-point-cloud-registration}, including the following:
\begin{itemize}
    \item For classical feature-free method, we choose \textit{Generalized ICP (GICP)} \cite{segalGeneralizedicp2009} due to its demonstrated advantage in bathymetric point cloud registration \cite{torrobaComparisonSubmapRegistration2018}.
     \item For classical feature-based method, we choose \textit{Fast Point Feature Histograms (FPFH)} \cite{rusuFastPointFeature2009}, a commonly used hand-crafted point cloud descriptor \cite{choyFullyConvolutionalGeometric2019, choyDeepGlobalRegistration2020, baiPointdscRobustPoint2021}.
    \item For learning-based dense feature descriptor, we choose \textit{Fully Convolutional Geometric Features (FCGF)} \cite{choyFullyConvolutionalGeometric2019},  the first fully convolutional point cloud feature extractor.
    \item For learning-based keypoint descriptor, we choose \textit{Predator} \cite{huangPredatorRegistration3d2021}, a method specialized at low overlap point cloud registration, since low overlap is a common condition for bathymetric survey data.
    \item For learning-based outlier rejection, we choose \textit{Deep Global Registration (DGR)} \cite{choyDeepGlobalRegistration2020} coupled with features learned by FCGF \cite{choyFullyConvolutionalGeometric2019}.
    \item Finally, we test \textit{BathyNN} \cite{tanDatadrivenLoopClosure2022}, a recent learning-based architecture designed specifically for loop closure detection of multibeam point clouds.
\end{itemize}
For feature-based methods, the predicted transformation is obtained using RANSAC with 50k iterations. For the number of points used during a RANSAC iteration, we found 3 to be optimal for \textit{FPFH}, whilst 4 yielded better performance for learning-based descriptors. For learning-based methods, we evaluate both pretrained models (KITTI pretrained models for \textit{FCGF}, \textit{DGR} and \textit{Predator}, and Antarctic 2019 dataset for \textit{BathyNN}) and models trained with our training set. Specifically, we are interested in whether models pretrained on similar metric scale data can be used off-the-shelf for the \mbesdataset. This experiment will help us better evaluate the uniqueness and challenges of our dataset.

\subsection{Implementation Details}
All learning-based methods were trained using submap pairs with all overlap ratios ($20\% - 100\%)$, or effective overlap ratios $(10\% - 50\%)$, resulting in a total number of 29046 training pairs. For each method, some hyperparameter search around the original paper's training setting was performed using a small subset of the training data.

For \textit{FCGF}, we finetuned from the KITTI pretrained model for 100 epochs using the Adam optimizer \cite{kingmaAdamMethodStochastic2017} with a OneCycle learning rate scheduler \cite{smith2019super} with maximum learning rate of 0.01. The batch size was set to 4, and hardest contrastive loss with positive margin 0.4 and negative margin 3.0 was used as the loss function. We then used this \textit{FCGF (trained)} model as feature extractor for \textit{DGR}, and trained the inlier classifier module in DGR. The \textit{DGR (trained)} model was trained for 20 epochs with a batch size of 8 using the ADAM optimizer with an Exponential learning rate schedule. The initial learning rate was 0.001, with a learning rate decay of $\gamma = 0.9$. For \textit{Predator}, we found finetuning from the KITTI model to be infeasible, and instead trained it from random initialization. \textit{Predator (trained)} was trained for 10 epochs with a batch size of 1 using the Stochastic Gradient Descent (SGD). Exponential learning rate scheduler with an initial learning rate of 0.05 and $\gamma = 0.95$ was used for training. This model has many dataset-specific hyperparameters. We set the number of negative pairs in circle loss $n_p = 256$, the temperature factor $\gamma = 24$, the voxel size $V= 1m$, the search radius for positive pairs $r_p = 1.5m$, safe rardius $r_s = 4.5m$, overlap radius $r_o = 2.0m$ and matchability radius $r_m = 2.0m$. The rest of model configurations follow the pretrained KITTI model. Finally, the \textit{BathyNN (trained)} is finetuned from the released Thwaites pretrained weights for 10 epochs, using the same hyperparameters as the original paper.

\subsection{Results}
We evaluate all methods on the test set of \mbesdataset. Noting that real-life AUV bathymetric surveys often generate data with low overlap, we separate the evaluation into different effective overlap ratios. For consistency error and transformation errors, we also report the values for ground truth transformation and null transformation (identity).

\subsubsection{Consistency Error}
\begin{figure}[h!]
    \centering
    \includegraphics[width=\linewidth]{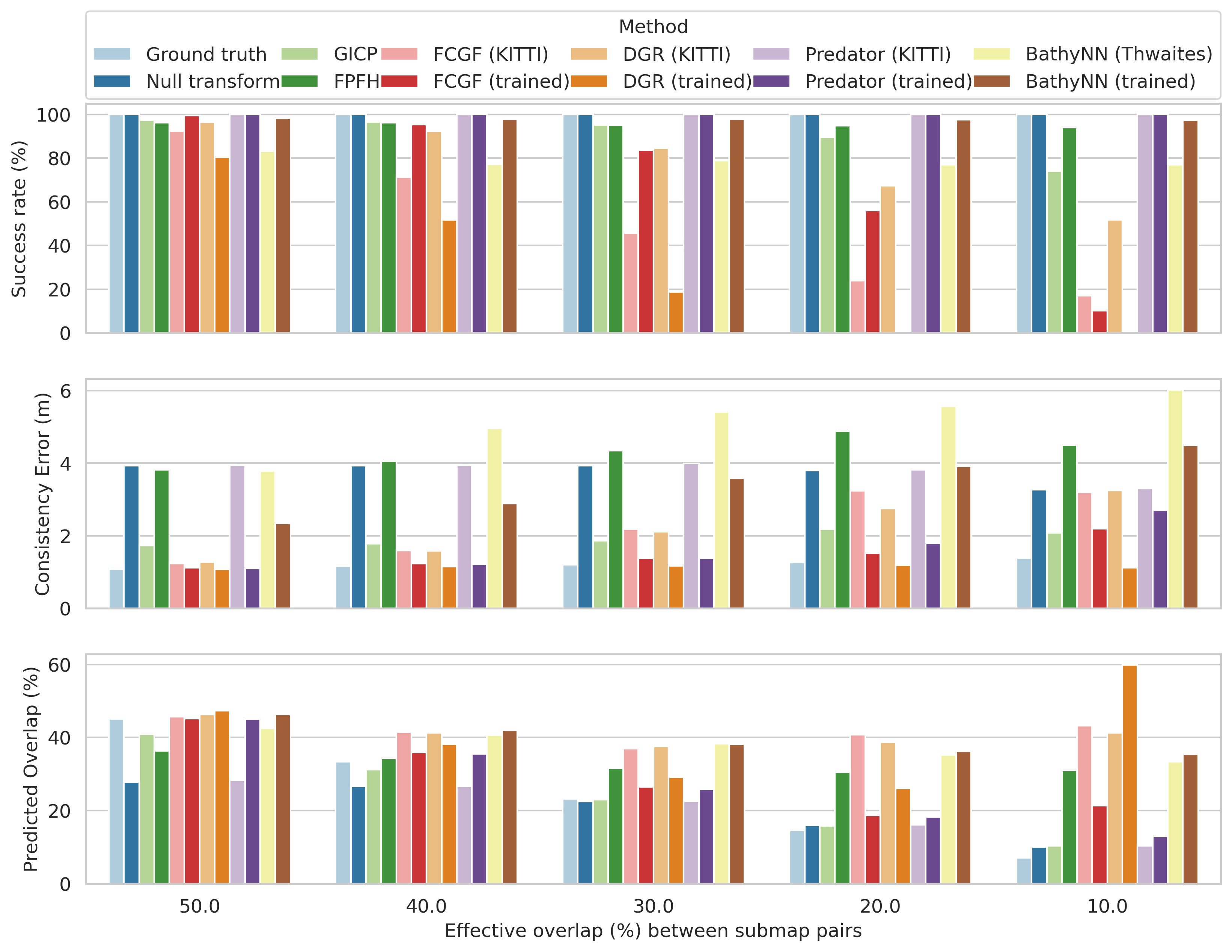}
    \caption{Map consistency metrics for decreasing overlap ratios. Note that the consistency and predicted overlap (\%) are only computed for successful registrations. Specifically, the success rate (\%) reported here is the rate at which the method returns a transform, regardless of its correctness. This is consistent with real-life AUV missions, where ground truth transformation is unknown. Transformation accuracy is evaluated in the subsequent metrics.}
    \label{fig:consistency}
\end{figure}

Looking at \autoref{fig:consistency}, we see that the KITTI models for FCGF and DGR can be readily applied to our dataset at high-overlap (50\%), though the performance quickly deteriorates as the overlap decreases. For the trained models, we see that Predator achieves a reasonable consistency even with an exceptionally high success rate. The trained FCGF and DGR trade a significant amount of success for a somewhat better map consistency. The BathyNN models specifically designed for MBES data are outperformed by all other neural learning-based models tested. GICP presents a reasonable out-of-the-box solution with a reasonable success rate as well as a reasonable consistency. In particular in low-overlap scenarios, GICP presents a robust solution. In higher-overlap scenarios, the deep learning methods can provide higher accuracy.

\subsubsection{Transformation Accuracy Metrics}
\begin{figure}[h!]
    \centering
    \includegraphics[width=\linewidth]{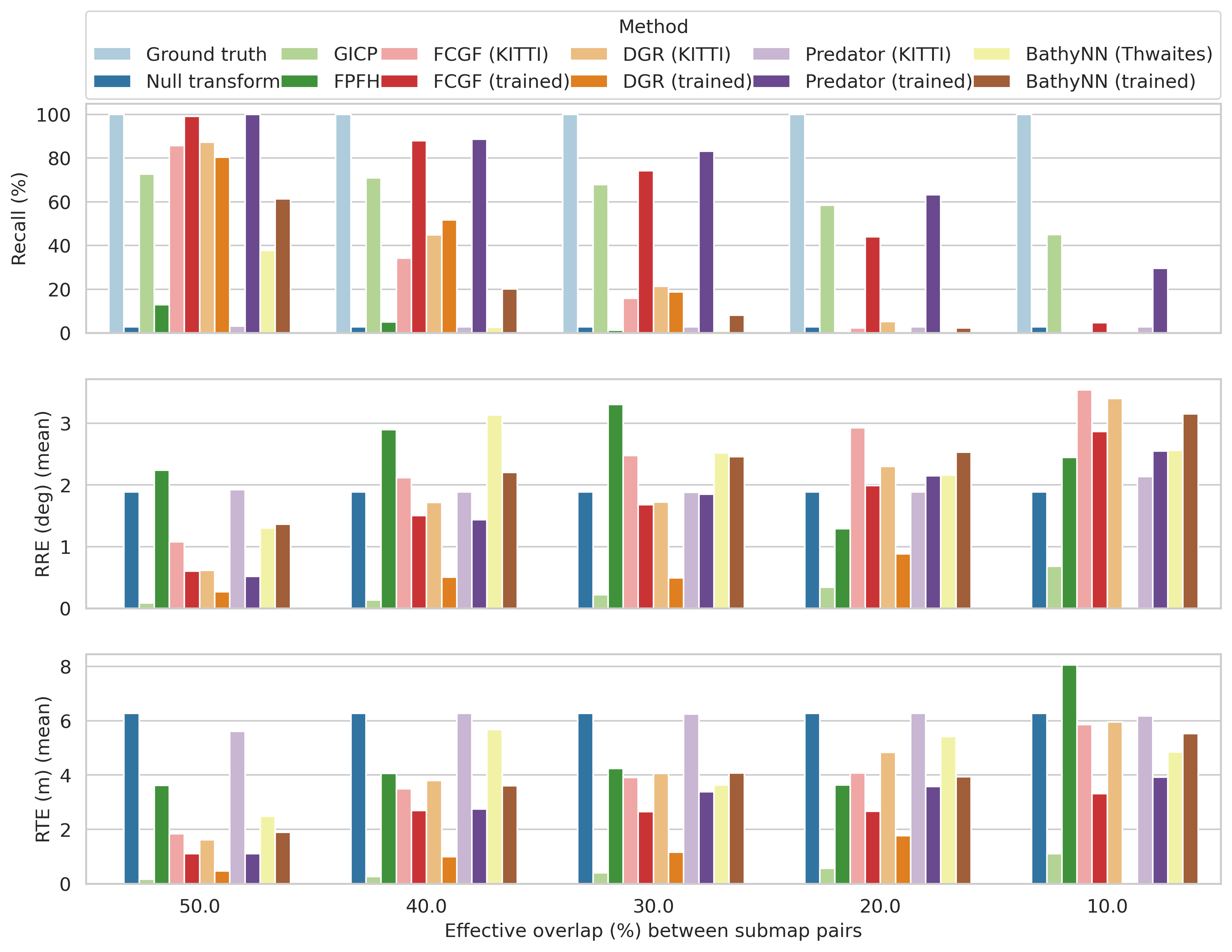}
    \caption{Registration error metrics given decreasing overlap ratios. Benefiting from the synthetic ground truth transformation, the recall (\%) here accounts for the cases where the predicted transformation is has an $RRE < 5 \degree$ and $RTE < 10m$.
    The RRE and RTE are only computed for the correctly recalled pairs.}
    \label{fig:transformation}
\end{figure}

The registration errors in \autoref{fig:transformation} reveal significant differences in the scale at which the methods perform well. At high overlap ratios, the neural network methods, and FCGF and Predator in particular, have a high recall ratio compared to GICP. This indicates an ability to recover a rough transform consistently. However, when looking at the transformation errors in the correctly recalled cases, it becomes apparent that GICP is more adept at finding the precise transform. From the results, it seems clear that GICP performs better across all metrics in the scenarios with very low overlap.

Comparing the error metrics in \autoref{fig:transformation} and \autoref{fig:consistency}, we notice that a low consistency error does not necessarily imply an accurate transformation. Specifically, at high overlap ratio (50-20\%), the trained FCGF and Predator models both have lower consistency error than GICP, yet the correctly recalled transformations from GICP have lower mean RRE and RTE than both models. This could be explained by two factors. Firstly, the consistency error is computed at $2m$ grid size, or twice the point cloud resolution. As such, it is better at identifying the rough correctness of the alignment, but might not be able to signal whether the transformation is precise. Secondly, MBES bathymetry data is often relatively flat, and small errors in transformation might not significantly affect the map quality.

\subsubsection{Feature Correspondence Metrics}
\begin{figure}[h!]
    \centering
    \includegraphics[width=\linewidth]{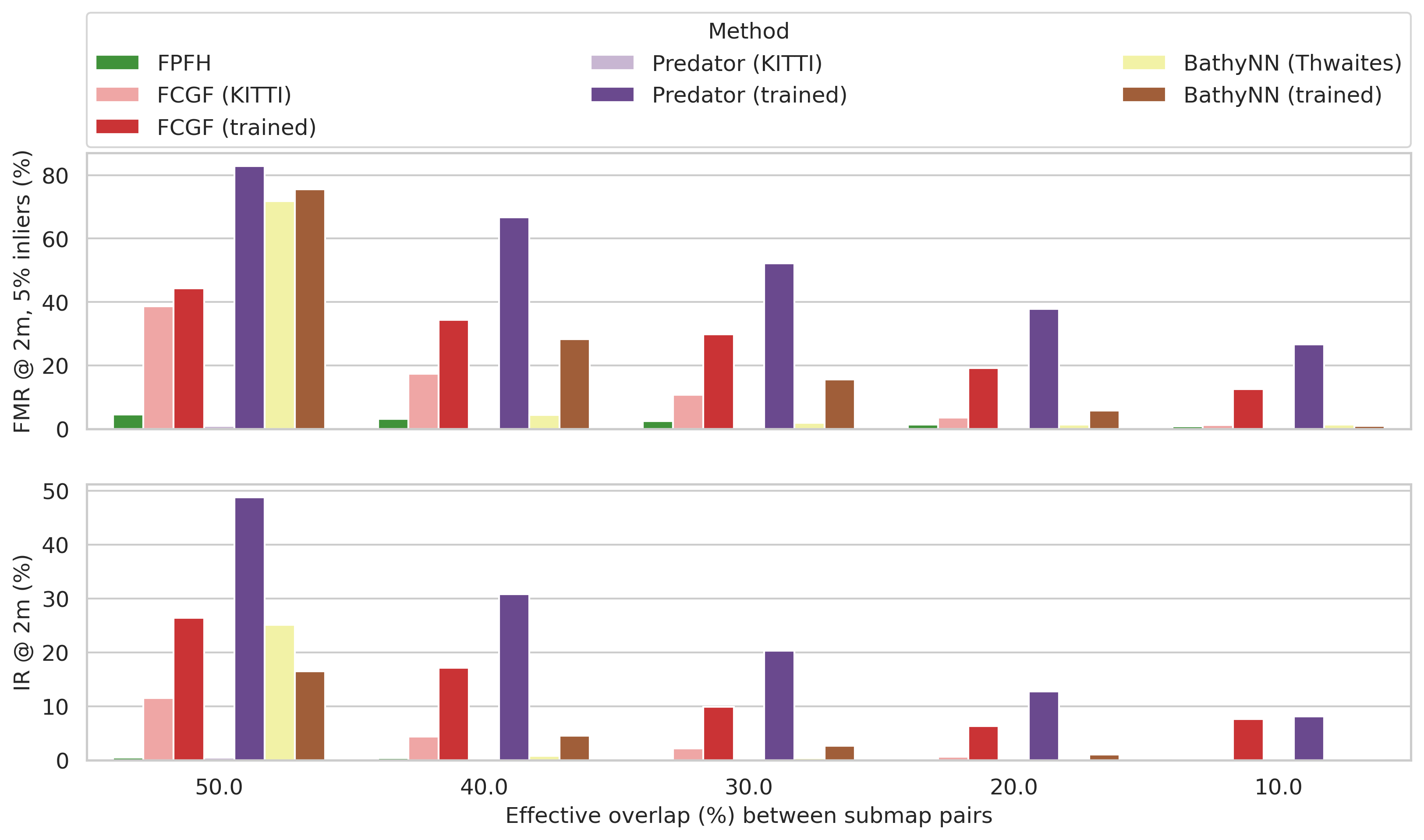}
    \caption{The feature match recall given decreasing overlap ratios. Note the poor performance of FPFH. The trained Predator clearly outperforms the other methods in this metric.}
    \label{fig:correspondence}
\end{figure}

From \autoref{fig:correspondence}, it is clear models finetuned on our dataset outperforms those trained purely on KITTI. 
This trend can also be observed in \autoref{fig:consistency} and \autoref{fig:transformation}. In fact, the best-performing method across all overlaps, Predator, barely finds any correct correspondences when trained only on KITTI. The significant increase in performance provided by fine-tuning clearly indicates the fundamental difference in registering Lidar data as compared to bathymetric sonar data.

\section{Conclusions}
\label{sec:sec5-conclusions}
In this paper, we introduced \mbesdataset~- a large-scale open source, semi-synthetic MBES registration dataset with a set of benchmark metrics. We see this work as a first step in constructing large scale MBES datasets that allows for easier benchmarking of point cloud registration methods. Using this dataset, we provide the first comprehensive evaluation of both classical and learning-based registration methods on bathymetric point cloud data. 
In general, we found learning-based methods, especially FCGF and Predator, to work well for initial coarse registration due to their high success rates. However, GICP provides the most robust registration, yielding precise transformations even at very low overlap (10\%). One natural extension of this work is thus applying a two-step registration for MBES-based SLAM problem, using Predator for coarse registration, and GICP for fine registration.

\newpage

\balance
\bibliographystyle{IEEEtran}
\bibliography{egbib}

\begin{thebibliography}{10}
\providecommand{\url}[1]{#1}
\csname url@samestyle\endcsname
\providecommand{\newblock}{\relax}
\providecommand{\bibinfo}[2]{#2}
\providecommand{\BIBentrySTDinterwordspacing}{\spaceskip=0pt\relax}
\providecommand{\BIBentryALTinterwordstretchfactor}{4}
\providecommand{\BIBentryALTinterwordspacing}{\spaceskip=\fontdimen2\font plus
\BIBentryALTinterwordstretchfactor\fontdimen3\font minus \fontdimen4\font\relax}
\providecommand{\BIBforeignlanguage}[2]{{%
\expandafter\ifx\csname l@#1\endcsname\relax
\typeout{** WARNING: IEEEtran.bst: No hyphenation pattern has been}%
\typeout{** loaded for the language `#1'. Using the pattern for}%
\typeout{** the default language instead.}%
\else
\language=\csname l@#1\endcsname
\fi
#2}}
\providecommand{\BIBdecl}{\relax}
\BIBdecl

\bibitem{micallefSubmarineGeomorphology2018}
A.~Micallef, S.~Krastel, and A.~Savini, Eds., \emph{Submarine {{Geomorphology}}}, ser. Springer {{Geology}}.\hskip 1em plus 0.5em minus 0.4em\relax {Cham}: {Springer International Publishing}, 2018.

\bibitem{graham2022rapid}
A.~G. Graham, A.~W{\aa}hlin, K.~A. Hogan, F.~O. Nitsche, K.~J. Heywood, R.~L. Totten, J.~A. Smith, C.-D. Hillenbrand, L.~M. Simkins, J.~B. Anderson \emph{et~al.}, ``Rapid retreat of thwaites glacier in the pre-satellite era,'' \emph{Nature Geoscience}, vol.~15, no.~9, pp. 706--713, 2022.

\bibitem{romanSelfConsistentBathymetricMapping2007}
C.~Roman and H.~Singh, ``A {{Self-Consistent Bathymetric Mapping Algorithm}},'' \emph{J. Field Robotics}, vol.~24, no. 1-2, pp. 23--50, Jan. 2007.

\bibitem{torrobaAutonomousIndustrialscaleBathymetric2019}
I.~Torroba, N.~Bore, and J.~Folkesson, ``Towards autonomous industrial-scale bathymetric surveying,'' in \emph{2019 {{IEEE}}/{{RSJ International Conference}} on {{Intelligent Robots}} and {{Systems}} ({{IROS}})}.\hskip 1em plus 0.5em minus 0.4em\relax {IEEE}, 2019, pp. 6377--6382.

\bibitem{torrobaComparisonSubmapRegistration2018}
------, ``A comparison of submap registration methods for multibeam bathymetric mapping,'' in \emph{2018 {{IEEE}}/{{OES Autonomous Underwater Vehicle Workshop}} ({{AUV}})}.\hskip 1em plus 0.5em minus 0.4em\relax {IEEE}, 2018, pp. 1--6.

\bibitem{tanDatadrivenLoopClosure2022}
J.~Tan, I.~Torroba, Y.~Xie, and J.~Folkesson, ``Data-driven loop closure detection in bathymetric point clouds for underwater slam,'' in \emph{2023 IEEE International Conference on Robotics and Automation (ICRA)}.\hskip 1em plus 0.5em minus 0.4em\relax IEEE, 2023, pp. 3131--3137.

\bibitem{romanConsistencyBasedError2006}
C.~Roman and H.~Singh, ``Consistency based error evaluation for deep sea bathymetric mapping with robotic vehicles,'' in \emph{2006 IEEE International Conference on Robotics and Automation (ICRA)}, May 2006, pp. 3568--3574.

\bibitem{brossardNewApproach3D2020}
M.~Brossard, S.~Bonnabel, and A.~Barrau, ``A {{New Approach}} to {{3D ICP Covariance Estimation}},'' \emph{IEEE Robotics and Automation Letters}, vol.~5, no.~2, pp. 744--751, Apr. 2020.

\bibitem{van2008visualizing}
L.~Van~der Maaten and G.~Hinton, ``Visualizing data using t-sne.'' \emph{Journal of machine learning research}, vol.~9, no.~11, 2008.

\bibitem{wu3dShapenetsDeep2015}
Z.~Wu, S.~Song, A.~Khosla, F.~Yu, L.~Zhang, X.~Tang, and J.~Xiao, ``3d shapenets: {{A}} deep representation for volumetric shapes,'' in \emph{Proceedings of the {{IEEE}} Conference on Computer Vision and Pattern Recognition}, 2015, pp. 1912--1920.

\bibitem{geigerAreWeReady2012}
A.~Geiger, P.~Lenz, and R.~Urtasun, ``Are we ready for autonomous driving? the {KITTI} vision benchmark suite,'' in \emph{2012 {{IEEE}} Conference on Computer Vision and Pattern Recognition}.\hskip 1em plus 0.5em minus 0.4em\relax {IEEE}, 2012, pp. 3354--3361.

\bibitem{zeng3dmatchLearningLocal2017}
A.~Zeng, S.~Song, M.~Nie{\ss}ner, M.~Fisher, J.~Xiao, and T.~Funkhouser, ``3dmatch: {{Learning}} local geometric descriptors from rgb-d reconstructions,'' in \emph{Proceedings of the {{IEEE}} Conference on Computer Vision and Pattern Recognition}, 2017, pp. 1802--1811.

\bibitem{barkbyFeaturelessApproachEfficient2011}
S.~Barkby, S.~B. Williams, O.~Pizarro, and M.~V. Jakuba, ``A featureless approach to efficient bathymetric {{SLAM}} using distributed particle mapping,'' \emph{Journal of Field Robotics}, vol.~28, no.~1, pp. 19--39, 2011.

\bibitem{yewRpmnetRobustPoint2020}
Z.~J. Yew and G.~H. Lee, ``Rpm-net: {{Robust}} point matching using learned features,'' in \emph{Proceedings of the {{IEEE}}/{{CVF}} Conference on Computer Vision and Pattern Recognition}, 2020, pp. 11\,824--11\,833.

\bibitem{huangComprehensiveSurveyPoint2021}
X.~Huang, G.~Mei, J.~Zhang, and R.~Abbas, ``A comprehensive survey on point cloud registration,'' \emph{arXiv preprint arXiv:2103.02690}, 2021.

\bibitem{zhangDeepLearningBased2020}
Z.~Zhang, Y.~Dai, and J.~Sun, ``Deep learning based point cloud registration: An overview,'' \emph{Virtual Reality \& Intelligent Hardware}, vol.~2, no.~3, pp. 222--246, 2020.

\bibitem{beslMethodRegistration3D1992}
P.~J. Besl and N.~D. McKay, ``Method for registration of 3-{{D}} shapes,'' in \emph{Sensor Fusion {{IV}}: Control Paradigms and Data Structures}, vol. 1611.\hskip 1em plus 0.5em minus 0.4em\relax {Spie}, 1992, pp. 586--606.

\bibitem{pomerleauReviewPointCloud2015}
F.~Pomerleau, F.~Colas, and R.~Siegwart, ``A review of point cloud registration algorithms for mobile robotics,'' \emph{Foundations and Trends\textregistered{} in Robotics}, vol.~4, no.~1, pp. 1--104, 2015.

\bibitem{fischlerRandomSampleConsensus1981}
M.~A. Fischler and R.~C. Bolles, ``Random sample consensus: A paradigm for model fitting with applications to image analysis and automated cartography,'' \emph{Communications of the ACM}, vol.~24, no.~6, pp. 381--395, 1981.

\bibitem{johnsonUsingSpinImages1999}
A.~E. Johnson and M.~Hebert, ``Using spin images for efficient object recognition in cluttered {{3D}} scenes,'' \emph{IEEE Transactions on pattern analysis and machine intelligence}, vol.~21, no.~5, pp. 433--449, 1999.

\bibitem{rusuFastPointFeature2009}
R.~B. Rusu, N.~Blodow, and M.~Beetz, ``Fast point feature histograms ({{FPFH}}) for {{3D}} registration,'' in \emph{2009 {{IEEE}} International Conference on Robotics and Automation (ICRA)}.\hskip 1em plus 0.5em minus 0.4em\relax {IEEE}, 2009, pp. 3212--3217.

\bibitem{tombariUniqueSignaturesHistograms2010}
F.~Tombari, S.~Salti, and L.~Di~Stefano, ``Unique signatures of histograms for local surface description,'' in \emph{Computer {{Vision}}\textendash{{ECCV}} 2010: 11th {{European Conference}} on {{Computer Vision}}, {{Heraklion}}, {{Crete}}, {{Greece}}, {{September}} 5-11, 2010, {{Proceedings}}, {{Part III}} 11}.\hskip 1em plus 0.5em minus 0.4em\relax {Springer}, 2010, pp. 356--369.

\bibitem{guoComprehensivePerformanceEvaluation2016}
Y.~Guo, M.~Bennamoun, F.~Sohel, M.~Lu, J.~Wan, and N.~M. Kwok, ``A comprehensive performance evaluation of {{3D}} local feature descriptors,'' \emph{International Journal of Computer Vision}, vol. 116, pp. 66--89, 2016.

\bibitem{choyFullyConvolutionalGeometric2019}
C.~Choy, J.~Park, and V.~Koltun, ``Fully convolutional geometric features,'' in \emph{Proceedings of the {{IEEE}}/{{CVF}} International Conference on Computer Vision}, 2019, pp. 8958--8966.

\bibitem{choy4dSpatiotemporalConvnets2019}
C.~Choy, J.~Gwak, and S.~Savarese, ``4d spatio-temporal convnets: {{Minkowski}} convolutional neural networks,'' in \emph{Proceedings of the {{IEEE}}/{{CVF}} Conference on Computer Vision and Pattern Recognition}, 2019, pp. 3075--3084.

\bibitem{huangPredatorRegistration3d2021}
S.~Huang, Z.~Gojcic, M.~Usvyatsov, A.~Wieser, and K.~Schindler, ``Predator: {{Registration}} of 3d point clouds with low overlap,'' in \emph{Proceedings of the {{IEEE}}/{{CVF Conference}} on Computer Vision and Pattern Recognition}, 2021, pp. 4267--4276.

\bibitem{zhouFastGlobalRegistration2016}
Q.-Y. Zhou, J.~Park, and V.~Koltun, ``Fast {{Global Registration}},'' in \emph{Computer {{Vision}} – {{ECCV}} 2016}, ser. Lecture {{Notes}} in {{Computer Science}}, B.~Leibe, J.~Matas, N.~Sebe, and M.~Welling, Eds.\hskip 1em plus 0.5em minus 0.4em\relax {Springer International Publishing}, pp. 766--782.

\bibitem{yangTeaserFastCertifiable2020b}
H.~Yang, J.~Shi, and L.~Carlone, ``Teaser: Fast and certifiable point cloud registration,'' \emph{IEEE Transactions on Robotics}, vol.~37, no.~2, pp. 314--333, 2020.

\bibitem{choyDeepGlobalRegistration2020}
C.~Choy, W.~Dong, and V.~Koltun, ``Deep global registration,'' in \emph{Proceedings of the {{IEEE}}/{{CVF}} Conference on Computer Vision and Pattern Recognition}, 2020, pp. 2514--2523.

\bibitem{baiPointdscRobustPoint2021}
X.~Bai, Z.~Luo, L.~Zhou, H.~Chen, L.~Li, Z.~Hu, H.~Fu, and C.-L. Tai, ``Pointdsc: {{Robust}} point cloud registration using deep spatial consistency,'' in \emph{Proceedings of the {{IEEE}}/{{CVF Conference}} on {{Computer Vision}} and {{Pattern Recognition}}}, 2021, pp. 15\,859--15\,869.

\bibitem{pais3dregnetDeepNeural2020}
G.~D. Pais, S.~Ramalingam, V.~M. Govindu, J.~C. Nascimento, R.~Chellappa, and P.~Miraldo, ``3dregnet: {{A}} deep neural network for 3d point registration,'' in \emph{Proceedings of the {{IEEE}}/{{CVF}} Conference on Computer Vision and Pattern Recognition}, pp. 7193--7203.

\bibitem{palomerMultibeam3DUnderwater2016}
A.~Palomer, P.~Ridao, and D.~Ribas, ``Multibeam {{3D Underwater SLAM}} with {{Probabilistic Registration}},'' \emph{Sensors}, vol.~16, no.~4, p. 560, Apr. 2016.

\bibitem{hammondAutomatedPointCloud2015}
M.~Hammond, A.~Clark, A.~Mahajan, S.~Sharma, and S.~Rock, ``Automated point cloud correspondence detection for underwater mapping using {{AUVs}},'' in \emph{{{OCEANS}} 2015 - {{MTS}}/{{IEEE Washington}}}.\hskip 1em plus 0.5em minus 0.4em\relax {Washington, DC}: {IEEE}, Oct. 2015, pp. 1--7.

\bibitem{sureshActiveSLAMUsing2020}
S.~Suresh, P.~Sodhi, J.~G. Mangelson, D.~Wettergreen, and M.~Kaess, ``Active {{SLAM}} using {{3D Submap Saliency}} for {{Underwater Volumetric Exploration}},'' in \emph{2020 {{IEEE International Conference}} on {{Robotics}} and {{Automation}} ({{ICRA}})}.\hskip 1em plus 0.5em minus 0.4em\relax {Paris, France}: {IEEE}, May 2020, pp. 3132--3138.

\bibitem{bore2018sparse}
N.~Bore, I.~Torroba, and J.~Folkesson, ``Sparse gaussian process slam, storage and filtering for auv multibeam bathymetry,'' in \emph{2018 IEEE/OES Autonomous Underwater Vehicle Workshop (AUV)}.\hskip 1em plus 0.5em minus 0.4em\relax IEEE, 2018, pp. 1--6.

\bibitem{torroba2023online}
I.~Torroba, M.~Cella, A.~Ter{\'a}n, N.~Rolleberg, and J.~Folkesson, ``Online stochastic variational gaussian process mapping for large-scale bathymetric slam in real time,'' \emph{IEEE Robotics and Automation Letters}, 2023.

\bibitem{dengImagenetLargescaleHierarchical2009}
J.~Deng, W.~Dong, R.~Socher, L.-J. Li, K.~Li, and L.~{Fei-Fei}, ``Imagenet: {{A}} large-scale hierarchical image database,'' in \emph{2009 {{IEEE}} Conference on Computer Vision and Pattern Recognition}.\hskip 1em plus 0.5em minus 0.4em\relax {Ieee}, 2009, pp. 248--255.

\bibitem{krasnoskyBathymetricMappingSLAM2022}
K.~Krasnosky, C.~Roman, and D.~Casagrande, ``A bathymetric mapping and {{SLAM}} dataset with high-precision ground truth for marine robotics,'' \emph{The International Journal of Robotics Research}, vol.~41, no.~1, pp. 12--19, Jan. 2022.

\bibitem{bernardiAURORAMultisensorDataset2022}
M.~Bernardi, B.~Hosking, C.~Petrioli, B.~J. Bett, D.~Jones, V.~A. Huvenne, R.~Marlow, M.~Furlong, S.~McPhail, and A.~Munaf{\`o}, ``{{AURORA}}, a multi-sensor dataset for robotic ocean exploration,'' \emph{The International Journal of Robotics Research}, vol.~41, no.~5, pp. 461--469, 2022.

\bibitem{bichucherBathymetricFactorGraph2015}
\BIBentryALTinterwordspacing
V.~Bichucher, J.~M. Walls, P.~Ozog, K.~A. Skinner, and R.~M. Eustice, ``Bathymetric factor graph {{SLAM}} with sparse point cloud alignment,'' in \emph{{{OCEANS}} 2015 - {{MTS}}/{{IEEE Washington}}}.\hskip 1em plus 0.5em minus 0.4em\relax {IEEE}, pp. 1--7. [Online]. Available: \url{http://ieeexplore.ieee.org/document/7404433/}
\BIBentrySTDinterwordspacing

\bibitem{wangDeepClosestPoint2019}
Y.~Wang and J.~M. Solomon, ``Deep closest point: {{Learning}} representations for point cloud registration,'' in \emph{Proceedings of the {{IEEE}}/{{CVF}} International Conference on Computer Vision}, 2019, pp. 3523--3532.

\bibitem{baiD3featJointLearning2020}
X.~Bai, Z.~Luo, L.~Zhou, H.~Fu, L.~Quan, and C.-L. Tai, ``D3feat: {{Joint}} learning of dense detection and description of 3d local features,'' in \emph{Proceedings of the {{IEEE}}/{{CVF Conference}} on {{Computer Vision}} and {{Pattern Recognition}}}, 2020, pp. 6359--6367.

\bibitem{segalGeneralizedicp2009}
A.~Segal, D.~Haehnel, and S.~Thrun, ``Generalized-icp.'' in \emph{Robotics: Science and Systems}, vol.~2, no.~4.\hskip 1em plus 0.5em minus 0.4em\relax {Seattle, WA}, p. 435.

\bibitem{kingmaAdamMethodStochastic2017}
D.~P. Kingma and J.~Ba, ``Adam: A method for stochastic optimization,'' \emph{arXiv preprint arXiv:1412.6980}, 2014.

\bibitem{smith2019super}
L.~N. Smith and N.~Topin, ``Super-convergence: Very fast training of neural networks using large learning rates,'' in \emph{Artificial intelligence and machine learning for multi-domain operations applications}, vol. 11006.\hskip 1em plus 0.5em minus 0.4em\relax SPIE, 2019, pp. 369--386.

\end{thebibliography}

\end{document}